\documentclass[pdflatex,sn-mathphys-num]{sn-jnl}

\usepackage{graphicx}%
\usepackage{multirow}%
\usepackage{amsmath,amssymb,amsfonts}%
\usepackage{amsthm}%
\usepackage{mathrsfs}%
\usepackage[title]{appendix}%
\usepackage{xcolor}%
\usepackage{textcomp}%
\usepackage{manyfoot}%
\usepackage{booktabs}%
\usepackage{algorithm}%
\usepackage{algorithmicx}%
\usepackage{algpseudocode}%
\usepackage{listings}%
\usepackage{gensymb}  

\theoremstyle{thmstyleone}%
\theoremstyle{thmstyletwo}%

\theoremstyle{thmstylethree}%

\usepackage{hyperref} 

\begin{document}

\title[RS2AD-LiDAR]{RS2AD-LiDAR: End-to-End Autonomous Driving LiDAR Data Generation from Roadside Sensor Observations}


\author[1,2,3]{\fnm{Runyi} \sur{Huang}}\email{hry21@mails.tsinghua.edu.cn}

\author[1,2,4]{\fnm{Ni} \sur{Ding}}\email{niding00@gmail.com}

\author[1,2,5]{\fnm{Ruidan} \sur{Xing}}\email{xingruidan2001@163.com}

\author[1,2]{\fnm{Yuheng} \sur{Shi}}\email{h3ngyu@gmail.com}

\author*[1,2]{\fnm{Lei} \sur{He}}\email{helei2023@tsinghua.edu.cn}

\author[1,2]{\fnm{Keqiang} \sur{Li}}\email{likq@tsinghua.edu.cn}

\affil*[1]{\orgdiv{State Key Laboratory of Intelligent Green Vehicle and Mobility}, \orgname{Tsinghua University}, \orgaddress{\city{Beijing}, \postcode{100084}, \country{China}}}

\affil*[2]{\orgdiv{School of Vehicle and Mobility}, \orgname{Tsinghua University}, \orgaddress{\city{Beijing}, \postcode{100084}, \country{China}}}

\affil[3]{\orgdiv{College of Artificial Intelligence}, \orgname{Tsinghua University}, \orgaddress{\city{Beijing}, \postcode{100083}, \country{China}}}

\affil[4]{\orgname{Logic \& Silicon AI Studio}}

\affil[5]{\orgdiv{School of Instrumentation and Optoelectronic Engineering}, \orgname{Beihang University}, \orgaddress{\city{Beijing}, \country{China}}}

\abstract{
End-to-end autonomous driving solutions, which directly process multimodal sensory data and output fine-grained control commands, have gradually become a mainstream direction with the development of autonomous driving technology. However, current methods in this category rely on single-vehicle data collection for model training and optimization, which suffers from high acquisition and annotation costs, scarcity of valuable scenarios, and data silos. To address these challenges, we propose RS2AD-LiDAR, a novel framework for reconstructing and generating vehicle-mounted LiDAR data from roadside sensor observations. 
Since no public dataset currently provides highly overlapping perception coverage between roadside and vehicle-mounted LiDAR sensors, which is essential for studying roadside-to-vehicle data generation, we constructed a dedicated dataset named R2V-LiDAR which is used solely for evaluation in this work.
Specifically, our method transforms roadside LiDAR point clouds into the vehicle-mounted LiDAR coordinate system, and synthesizes high-fidelity vehicle-mounted data via virtual LiDAR modeling and point cloud resampling techniques.
To the best of our knowledge, this is the first approach to reconstruct vehicle-mounted LiDAR data from roadside sensor inputs. Extensive experimental comparisons demonstrate the semantic similarity between the generated data and real data. Furthermore, object detection experiments show that incorporating the generated data into real data for model training improves both Bird's Eye View (BEV) and 3D detection accuracy, thereby validating the effectiveness of the proposed method.}

\keywords{Autonomous Driving, LiDAR Data Generation, End-to-End System, Object Detection}



\maketitle

\begin{figure}[htbp]
    \centering
    \includegraphics[width=0.9\textwidth]{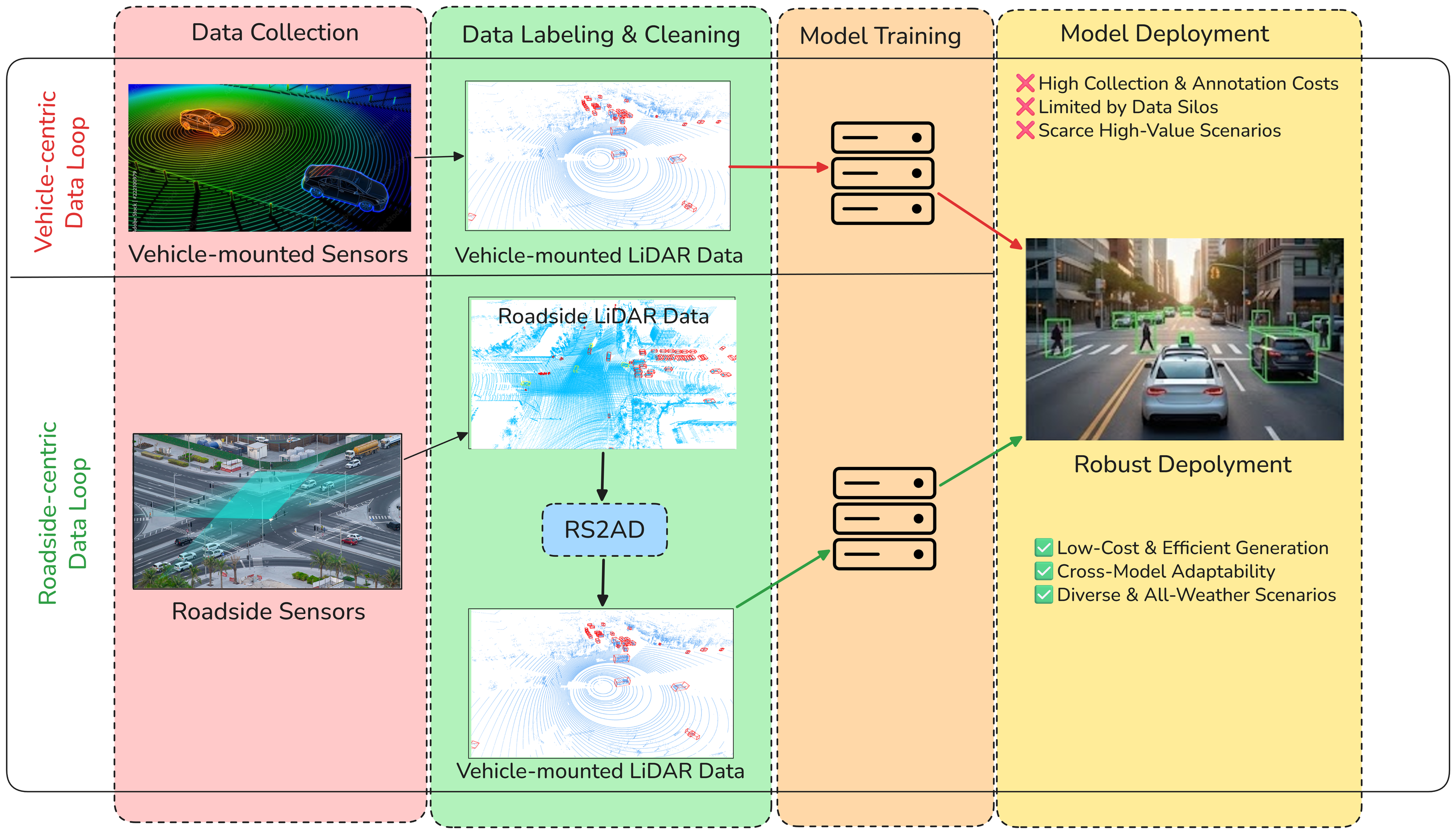}
    \caption{Comparison between the conventional vehicle-centric data loop and the proposed roadside-infrastructure-centric data loop. Our approach leverages the advantages of roadside sensors for long-term, real-world traffic data collection. This data is then used to generate synthetic vehicle-side data for model training.}
    \label{fig1}
\end{figure}

\section{Introduction}\label{sec1}

End-to-end autonomous driving models, exemplified by Tesla FSD, aim to fully neuralize core autonomous driving algorithms \cite{chen2024endtoend}, significantly reducing the reliance on manually coded rule-based logic. These models have gradually become the mainstream trend in autonomous driving technology \cite{wang2024drive, lu2025hierarchical}. However, current end-to-end autonomous driving systems still face several challenges, including high data collection and annotation costs, scarcity of high-value driving scenarios, and difficulties in effectively mining such scenarios \cite{pitropov2021canadian}. Additionally, data silos arise as individual automakers restrict data loops to their own vehicle models, limiting cross-vehicle and cross-scenario adaptability and hindering model generalization.

The above challenges largely stem from the inherent limitations of vehicle-centric data collection paradigms. As shown in Fig. \ref{fig1}, current autonomous driving data loops primarily rely on large-scale fleet deployment, where data is collected from individual vehicles, followed by data cleaning, annotation, and then used for model training. This paradigm leads to high data acquisition and annotation costs, limited coverage of long-tail traffic scenarios, and restricted cross-vehicle generalization because data distributions are tightly coupled with specific vehicle platforms and sensor configurations. Moreover, simulation-based data generation methods often suffer from Sim2Real domain gaps due to discrepancies between simulated and real-world traffic distributions.

To address these limitations, we propose a novel method to reconstruct vehicle-mounted LiDAR data from roadside sensor data, leveraging roadside sensors to generate high-precision dynamic and static 3D scene reconstructions. Roadside sensors naturally observe dense and continuous real-world traffic flows, providing realistic scene distributions and long-term stable observations. 
In addition, roadside sensing is largely free from vehicle height constraints and severe self-occlusion effects, enabling the acquisition of more comprehensive point cloud information  \cite{gao2024vehicle}.
By directly leveraging real-world roadside traffic distributions, our method inherently mitigates the Sim2Real gap. Furthermore, through virtual LiDAR modeling and flexible sampling strategies, the generated data can be adapted to arbitrary vehicle models and LiDAR configurations, enabling scalable cross-vehicle data reuse while significantly reducing data annotation costs. This also makes it possible to substantially expand the scale and diversity of training datasets without incurring additional onboard data collection costs, particularly for rare edge cases that are difficult to capture from the vehicle perspective.

Specifically, our method reconstructs vehicle-mounted LiDAR measurements by transforming roadside LiDAR point clouds into the target vehicle coordinate system using relative pose estimation. Virtual LiDAR sampling is then performed using geometry-aware plane fitting strategies to ensure both structural consistency and sampling accuracy, enabling high-fidelity reconstruction of both static and dynamic scene elements. Unlike traditional vehicle-centric data loops, our approach enables infrastructure-centric scalable data generation with real-world traffic distribution consistency and flexible sensor adaptation capability. This paradigm enables a shift from vehicle-centric data scaling to infrastructure-assisted data scaling, where data collected once at roadside infrastructure can be reused across multiple vehicle platforms, sensor configurations, and driving scenarios. To facilitate the study and evaluation of this infrastructure-
centric data loop, we construct a dedicated R2V-LiDAR dataset featuring highly overlapping perception coverage between roadside and vehicle-mounted LiDAR sensors.

The remainder of this paper is organized as follows. Section \ref{sec2} reviews related work on autonomous driving perception and data generation. Section \ref{sec3}  introduces the proposed RS2AD-LiDAR framework and details the vehicle-mounted LiDAR reconstruction methodology. Section \ref{sec4} presents experimental results and analyses on the R2V-LiDAR dataset. Finally, Section \ref{sec5} concludes this paper and discusses future research directions.

The main contributions of our work are as follows:

\begin{itemize}
    \item The first method to reconstruct vehicle-mounted LiDAR sensor data from roadside sensor inputs, providing a new low-cost and scalable paradigm for data generation in autonomous driving.
    
    \item A novel LiDAR sampling and modeling approach that mitigates the Sim2Real adaptation gap by improving data realism and consistency, while supporting arbitrary vehicle models and LiDAR sensor configurations for flexible adaptation to different beam settings.
    
    \item Comprehensive experiments conducted on our self-collected R2V-LiDAR dataset demonstrate the effectiveness of the proposed method and validate its capability to efficiently generate training data for end-to-end autonomous driving systems under diverse sensor configurations.
\end{itemize}

\section{Related Work}\label{sec2}

The rapid advancement of autonomous driving research is fundamentally driven by the availability of large-scale and high-quality datasets. However, the high cost associated with collecting point cloud data, together with severe data silos across vehicle platforms and sensor configurations, poses significant challenges for acquiring sufficient and diverse real-world data for deep learning-based autonomous driving systems. Consequently, scalable and efficient data acquisition and data expansion paradigms have become a key research focus in recent years.

Existing efforts can be broadly grouped into three main directions: (1) real-world autonomous driving datasets collected under different sensing paradigms, including vehicle-mounted, roadside infrastructure-based, and cooperative vehicle–infrastructure sensing datasets; (2) data generation and augmentation methods built upon real-world datasets; and (3) simulation-based data synthesis methods using virtual environments. 

\subsection{Real-World Autonomous Driving Datasets}
Acquiring datasets from real vehicles is the most intuitive approach, and the release of various large-scale autonomous driving datasets has significantly propelled research in this field. Existing datasets can be broadly categorized according to their sensing paradigms, including vehicle-mounted datasets, roadside infrastructure datasets, and cooperative vehicle–infrastructure datasets.

Vehicle-mounted datasets represent the most widely adopted data acquisition paradigm. Early benchmarks such as KITTI  \cite{geiger2012we} established standardized evaluation protocols for 3D perception tasks. Later Cityscapes \cite{cordts2016cityscapes} primarily provide 2D annotations for images, aiming for semantic segmentation. More recent large-scale datasets, including nuScenes  \cite{caesar2020nuscenes} and Waymo Open Dataset  \cite{sun2020scalability}, provide multi-modal sensor data including camera images and LiDAR point clouds, significantly advancing deep learning-based autonomous driving perception. 
These datasets are fundamentally collected under a vehicle-centric sensing paradigm, where data distributions are tightly coupled with specific vehicle platforms and driving trajectories. Moreover, vehicle-mounted sensing is inherently affected by self-occlusion and near-field occlusion from surrounding vehicles, pedestrians, and roadside objects, which limits the perception range and completeness of scene observations.

Compared with vehicle-mounted datasets, roadside infrastructure sensing datasets alleviate occlusion issues by providing higher and wider observation viewpoints that reduce near-field occlusions.
Rope3D \cite{ye2022rope3d} provides large-scale roadside perception benchmarks with diverse viewpoints and environmental conditions, enabling research on roadside 3D object detection from monocular camera observations.
To address the scarcity of roadside LiDAR datasets, the TUM Traffic (TUMTraf) Intersection Dataset \cite{zimmer2023tumtraf} provides more high-quality 3D annotated roadside LiDAR point clouds synchronized with camera images, contributing to complex 3D camera-LiDAR roadside perception tasks.
Pure roadside datasets are primarily designed for infrastructure perception and traffic monitoring, and typically lack explicit cross-platform data correspondence with vehicle-mounted observations.

Cooperative perception datasets further extend sensing coverage by combining vehicle-side and infrastructure-side observations. DAIR-V2X \cite{yu2022dairv2x} is the first large-scale real-world vehicle–infrastructure cooperative perception dataset, providing standardized benchmarks for cross-view and cross-agent perception learning. Similarly, HoloVIC \cite{ma2024holovic} constructs large-scale holographic vehicle–infrastructure cooperative datasets using multiple sensor layouts across complex intersections, providing synchronized multi-sensor data and cross-device object association annotations. While these datasets play a critical role in advancing cooperative perception research, they are primarily designed to enhance vehicle perception using infrastructure assistance, rather than enabling infrastructure-to-vehicle data generation. In particular, the sensing coverage between vehicle-mounted sensors and roadside sensors in these datasets does not necessarily exhibit strict field-of-view overlap, limiting their direct applicability to roadside-to-vehicle data reconstruction tasks.

\subsection{Data Generation Based on Real-World Autonomous Driving Data}

Direct data collection and annotation are costly and labor-intensive, and real-world datasets often fail to comprehensively cover rare and complex long-tail scenarios. Thus many studies attempt to expand training data through generation and augmentation techniques built upon existing onboard datasets. 

Early point cloud generation methods mainly rely on latent generative models. For example, Caccia et al. \cite{caccia2019deep} proposed a generative framework based on spherical range image representations using Generative Adversarial Networks (GANs) and Variational Autoencoders (VAEs), while LiDARGen \cite{zyrianov2022learning} further introduced diffusion-based generation in equirectangular range-view space. However, range-view representations project 3D point clouds onto regular 2D grids, which may introduce geometric distortions and quantization errors during 3D–2D projection \cite{kong2023rethinking}.

To address these representation limitations, alternative spatial encodings have been explored. UltraLiDAR \cite{xiong2023learning} adopts a BEV voxel grid representation to explicitly encode geometric structures and occlusion relationships. LiDARDM  \cite{zyrianov2025lidardm} further integrates latent diffusion models with physical simulation to generate temporally consistent LiDAR sequences.

Recently, large-scale generative world models, such as Cosmos-Drive-Dreams \cite{nvidia2025cosmosdrivedreams}, have demonstrated strong capabilities in generating high-fidelity LiDAR sequences conditioned on structured inputs such as HD maps or images. Despite their strong generation capability, these approaches typically require massive pretraining datasets and substantial computational resources, limiting their scalability and accessibility.

Overall, generation methods built upon real-world datasets significantly improve data diversity and learning robustness. However, most existing approaches inherit the sensing distributions and viewpoint constraints of the original vehicle-mounted data, making it difficult to achieve flexible cross-vehicle data reuse or sensor-configuration-level scalability.

\subsection{Data Synthesis Based on Environmental Simulation}

While real-world datasets provide physically accurate sensor measurements and realistic traffic distributions, their collection and annotation are expensive, time-consuming, and often limited in scenario diversity and long-tail event coverage. To overcome these limitations, researchers have increasingly explored simulation-based data synthesis, where large-scale labeled perception data can be generated in controllable virtual environments.

Simulation-based point cloud generation methods synthesize point clouds by simulating LiDAR measurements within virtual environments. Platforms such as CARLA \cite{dosovitskiy2017carla}, Gazebo \cite{koenig2004design}, and AirSim \cite{shah2017airsim} provide high-fidelity environments and configurable sensor models, enabling the generation of richly annotated synthetic point clouds. In addition to simulation platforms, synthetic autonomous driving datasets such as SYNTHIA \cite{ros2016synthia} further demonstrate the potential of virtual data generation by providing large-scale annotated synthetic driving scenes for perception model training and evaluation. However, physics-based simulations rely on manually crafted 3D assets, limiting scene diversity and accurately capturing dynamic real-world phenomena, resulting in a persistent sim-to-real gap. 

To mitigate this gap, data-driven approaches have been proposed: LiDARSim \cite{manivasagam2020lidarsim} corrects simulation errors via deep learning, while NeRF-based methods  \cite{huang2023neural}\cite{tao2024lidar} improve asset reconstruction quality to better approximate real point cloud distributions. Despite these advances, such methods remain dependent on virtual environments or high-quality reconstruction models, resulting in complex generation pipelines and substantial computational overhead. Overall, existing works either rely on simulation environments or vehicle-centric sensing paradigms, leaving roadside-to-vehicle reconstruction largely unexplored.

\section{Method}\label{sec3}

This section presents the technical implementation details of the RS2AD-LiDAR system. To enable the generation of vehicle-mounted LiDAR data from roadside observations, we construct the R2V-LiDAR dataset. This dataset is characterized by a highly overlapping field of view between roadside and vehicle-mounted sensors, and records continuous spatio-temporal observations as the ego vehicle passes through intersection regions. All experiments in this paper are conducted on this dataset. The parameters of the real vehicle-mounted LiDAR sensor on the ego vehicle (Hesai Pandar64 LiDAR) are used as the configuration baseline for the virtual LiDAR.

Based on this foundation, the proposed method can be divided into five core modules: target-view virtual LiDAR modeling, cross-view spatial alignment and perception region filtering, ground scene decomposition, point cloud resampling generation, and cross-domain label mapping module. Specifically, given a frame of point cloud data and corresponding annotations captured by a roadside sensor, a labeled vehicle in the scene is first selected as the target, and a virtual vehicle-mounted LiDAR model is established from the target perspective. Next, the cross-view spatial alignment and perception region filtering module utilizes coordinate transformation relationships to obtain the point cloud representation in the target vehicle LiDAR coordinate system, while removing points that fall outside the effective sensing range of the sensor. The ground scene decomposition module then applies the Patchwork++ \cite{lee2022patchwork++} method to separate the point cloud into ground and non-ground points. This enables the point cloud resampling generation module to adopt different planar modeling strategies and perform ray-casting-based resampling to generate target-view point clouds for ground and non-ground regions, respectively. Finally, the cross-domain label mapping module transforms roadside annotations into the target-view coordinate system using the established transformation relationships, thereby avoiding the additional cost of manual re-annotation.

\begin{figure}[h]
    \centering
    \includegraphics[width=0.9\textwidth]{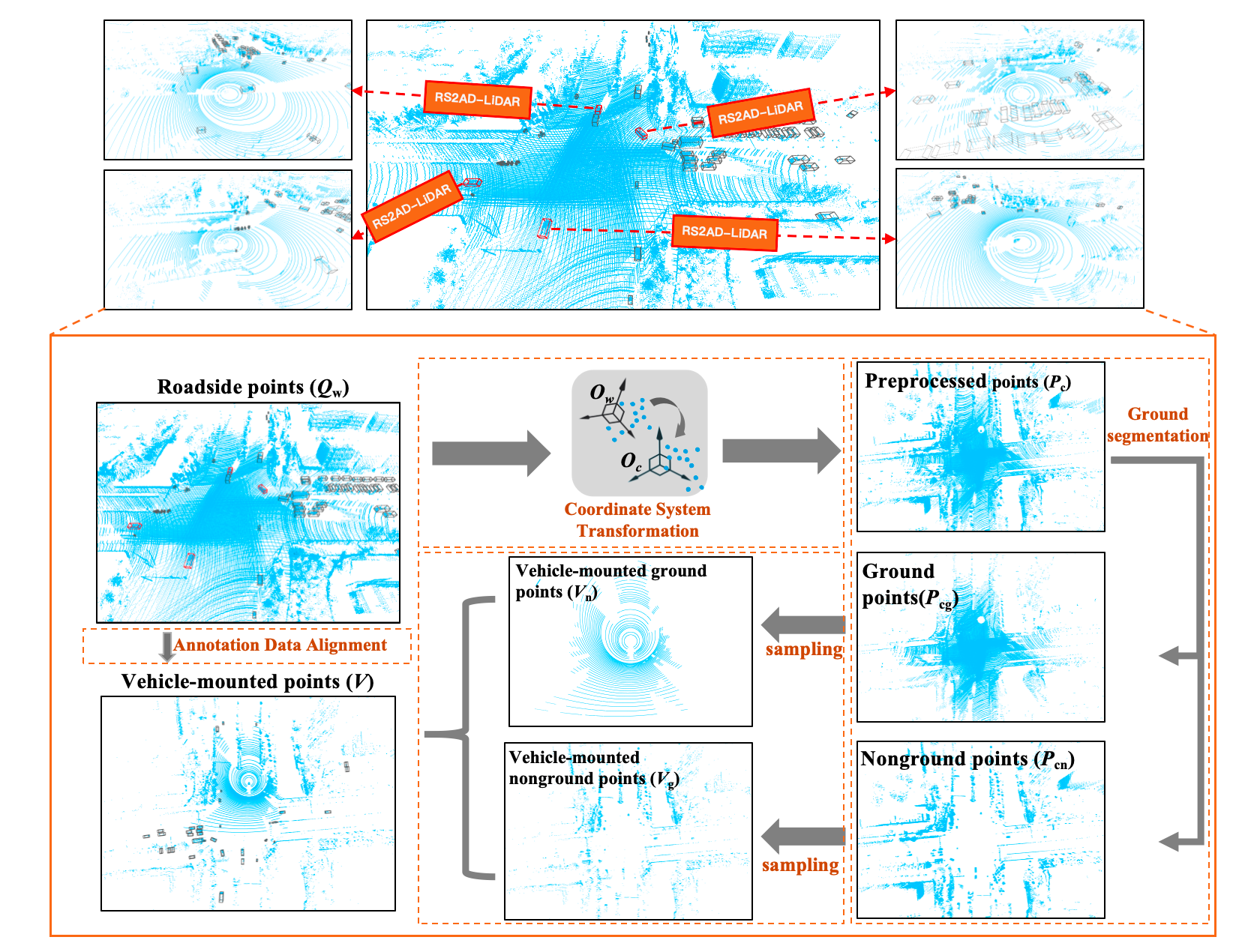}
    \caption{Overview of the RS2AD-LiDAR framework. The lower part depicts the system architecture, detailing the pipeline from roadside data acquisition to vehicle-mounted point cloud generation. The upper part provides a qualitative demonstration: the central panel shows the original roadside LiDAR scan, with the left and right panels displaying the reconstructed vehicle-mounted LiDAR data for four selected vehicles.}
    \label{fig2}
\end{figure}

Through these core components, for a single frame of roadside scan data, the RS2AD-LiDAR system can select any annotated vehicle in the scene as the target viewpoint and generate the corresponding vehicle-view point cloud. Global annotations can be transformed accordingly into the target coordinate system. The overall pipeline and generation results are illustrated in Fig. \ref{fig2}.

\subsection{Target-View Virtual LiDAR Modeling}\label{sec31}

After selecting a labeled vehicle of interest in the scene, a virtual vehicle-mounted LiDAR model and its corresponding coordinate system $O_l\!-\!x_ly_lz_l$ are first established. This model is designed to simulate the working mechanism of a real rotating LiDAR, and its default configuration is based on the Pandar64 LiDAR sensor. The virtual LiDAR emits laser beams while rotating around its \(z_l\)-axis, forming a set of fan-shaped scanning planes. These scanning planes are perpendicular to the horizontal plane $x_lO_ly_l$ and parallel to the $z_l$ axis. The configurable parameters of the model include the vertical field of view (FOV) $\alpha$, horizontal field of view $\beta$ (typically set to $360^\circ$), number of vertical scan lines $k$, the elevation angle distribution of each scan ray, horizontal angular resolution $\Delta \beta$, and the effective detection range $[r_{\min}, r_{\max}]$.

Here, the vertical field of view $\alpha$ represents the overall coverage range of the LiDAR in the vertical direction. According to the scan line configuration of the LiDAR sensor, the elevation angle of a single laser beam relative to the horizontal plane is denoted by $\epsilon$. To facilitate range filtering operations, this paper additionally introduces the spherical coordinate system and the polar angle $\theta$, defined as the angle between the ray direction and the positive direction of the $z_l$ axis. The relationship between the elevation angle $\epsilon$ and the polar angle $\theta$ is given by:

\begin{equation}
\theta = \frac{\pi}{2} - \epsilon
\label{eq:polar_angle}
\end{equation}

The virtual vehicle-mounted LiDAR emits rays outward from the origin of the coordinate system. These rays are represented as a set $\mathcal{L} = \{l_{ij}\}$. The vertical index $j=0, 1, 2, \dots, k-1$ corresponds to the $j$-th scan line, whose elevation angle is denoted as $\epsilon_j$ and is determined by the sensor calibration information. The horizontal index $i=0, 1, 2, \dots, m-1$ corresponds to the horizontal scanning angle, whose azimuth angle is defined as

\begin{equation}
\phi_i = i \cdot \Delta \beta
\label{eq:phi_angle}
\end{equation}

where $m = \beta / \Delta \beta$ is determined by the horizontal field of view and the horizontal angular resolution.

Under the spherical coordinate representation, the direction of each ray can be uniquely determined by the polar angle $\theta_j$ and the azimuth angle $\phi_i$. The unit direction vector of the virtual ray $l_{ij}$ can be expressed as

\begin{equation}
\mathbf{d}_{ij} =
\begin{bmatrix}
\sin\theta_j \cos\phi_i \\
\sin\theta_j \sin\phi_i \\
\cos\theta_j
\end{bmatrix}
\label{eq:ray_direction}
\end{equation}

For each virtual ray, the subsequent modules perform a geometry-constrained point cloud resampling process to estimate the corresponding LiDAR return point under the target viewpoint. Finally, all valid resampled points are aggregated to construct the point cloud data under the target vehicle viewpoint.

\subsection{Cross-View Spatial Alignment and Perception Region Filtering Module}

\begin{figure}[h]
    \centering
    \includegraphics[width=0.9\textwidth]{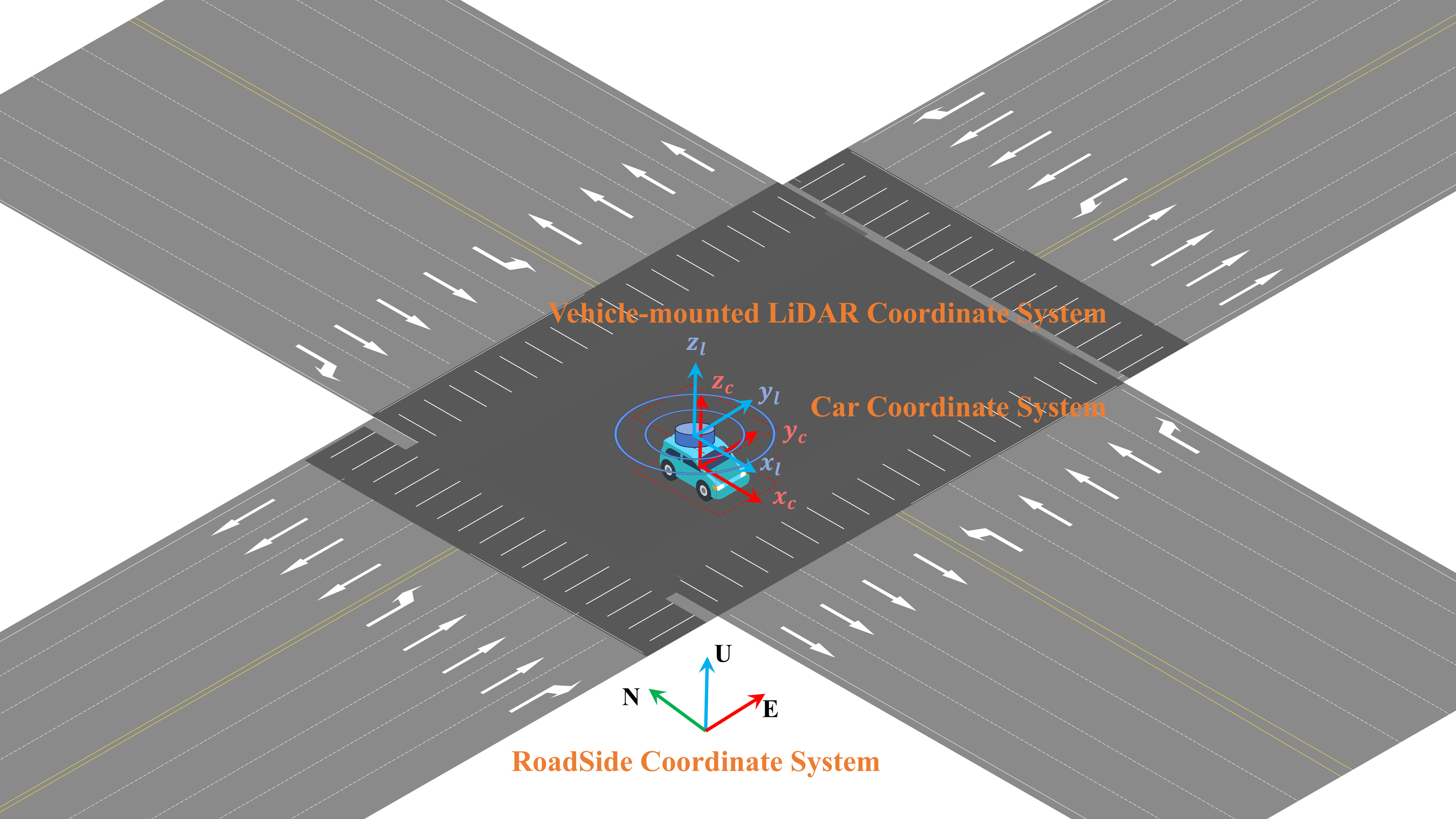}
    \caption{The coordinate systems in R2V-LiDAR dataset and RS2AD-LiDAR system, involving roadside world coordinate system, car coordinate system and vehicle-mounted LiDAR coordinate system.}
    \label{fig3}
\end{figure}

This module aims to achieve cross-view spatial alignment of roadside point cloud data to the target vehicle-mounted LiDAR coordinate system, and to perform effective filtering of the point cloud according to the sensing range of the virtual vehicle-mounted LiDAR, thereby reducing computational overhead caused by invalid points.

The data used in this study are collected from the self-constructed R2V-LiDAR dataset. This dataset contains cross-platform sensing data including roadside perception and vehicle-mounted perception. Therefore, it is necessary to describe multi-source sensor data under a unified spatial reference framework. To this end, a multi-layer coordinate system is constructed, including the roadside world coordinate system, the vehicle coordinate system, and the vehicle-mounted LiDAR coordinate system. All coordinate systems follow the right-hand rule. The definitions of each coordinate system are as follows, as illustrated in Fig. \ref{fig3}.

\textbf{Roadside World Coordinate System} adopts the East-North-Up (ENU) coordinate convention and is denoted as
$O_w\!-\!x_wy_wz_w$. In this coordinate system, the $x_w$ axis points to the east direction, the $y_w$ axis points to the north direction, and the $z_w$ axis points vertically upward. A stable reference point in the intersection scene is selected as the coordinate origin. The point cloud collected by the roadside LiDAR is denoted as
$Q_w=\{p_w^i\}_{i=1}^N$, where
$p_w^i=(x_w^i,y_w^i,z_w^i)^\top$ represents the 3D spatial position of the point in the roadside world coordinate system.

\textbf{Vehicle Coordinate System} is denoted as $O_c\!-\!x_cy_cz_c$, with the geometric center of the vehicle from roadside annotations defined as the coordinate origin. The axis directions follow the commonly used definition in autonomous driving: the $x_c$ axis points to the forward driving direction, the $y_c$ axis points to the left side of the vehicle, and the $z_c$ axis points vertically upward.

\textbf{Vehicle-Mounted LiDAR Coordinate System} is denoted in Cartesian form as $O_l\!-\!x_ly_lz_l$, whose axis directions are generally consistent with those of the vehicle coordinate system, and whose origin is located at the optical center of the LiDAR sensor. The LiDAR spherical coordinate system is built upon the Cartesian coordinate system of the LiDAR. In the LiDAR Cartesian coordinate system, any spatial point $(x, y, z)$ can be uniquely mapped to a spherical coordinate triplet $(r, \phi, \theta)$, where $r$ represents the radial distance (range) from the target point to the LiDAR origin, $\phi$ represents the azimuth angle of the target point relative to the $x$ axis in the $xy$ plane, and $\theta$ represents the polar angle measured from the positive $z$ axis direction. The conversion relationships are defined as follows:

\begin{equation}
r = \sqrt{x^2 + y^2 + z^2}
\label{eq:range_def}
\end{equation}
\begin{equation}
\phi = \mathrm{atan2}(y, x)
\label{eq:azimuth_def}
\end{equation}
\begin{equation}
\theta = \arccos\left(\frac{z}{r}\right)
\label{eq:polar_def}
\end{equation}

Cross-view spatial alignment requires transforming the roadside LiDAR point cloud data $Q$ from the roadside world coordinate representation $Q_w=\{p_w^i\}_{i=1}^N$ into the vehicle-mounted LiDAR coordinate representation $Q_l=\{p_l^i\}_{i=1}^N$. Therefore, it is necessary to obtain the homogeneous transformation matrix from the roadside world coordinate system to the vehicle-mounted LiDAR coordinate system:
\begin{equation}
T_{lw} =
\begin{bmatrix}
R_{lw}^{3 \times 3} & t_{lw}^{3 \times 1} \\
0 & 1
\end{bmatrix}
\end{equation}
where the left-multiplication column-vector convention is adopted. $R_{lw} \in \mathbb{R}^{3 \times 3}$ represents the rotation matrix from the roadside world coordinate system to the LiDAR coordinate system, and $t_{lw} \in \mathbb{R}^3$ represents the corresponding translation vector.

The acquisition of $T_{lw}$ requires using the vehicle coordinate system as an intermediate bridge. After selecting a labeled target vehicle of interest in the scene, the roadside annotation information contains the 3D coordinate of the vehicle center in the roadside world coordinate system $X_w$ and the vehicle rotation vector relative to the roadside world coordinate system $\Theta_w$, i.e., the translation vector $t_{wc} \in \mathbb{R}^3$ and rotation vector $r_{wc} \in \mathbb{R}^3$ from the vehicle coordinate system to the roadside world coordinate system. The rotation vector represents the 3D rotation relationship in Euler angle form and can be converted into a 3D rotation matrix $R_{wc}$ using the Rodrigues formula:
\begin{equation}
R_{wc} = \mathrm{Rodrigues}(r_{wc})
\label{eq:rodrigues}
\end{equation}

By applying the inverse operation, the transformation matrix from the roadside world coordinate system to the vehicle coordinate system $T_{cw}$ can be obtained:
\begin{equation}
    T_{cw} = T_{wc}^{-1} =
    \begin{bmatrix}
    R_{wc}^{-1} & -R_{wc}^{-1} \cdot t_{wc} \\
    0 & 1
    \end{bmatrix}
\end{equation}

According to the sensor extrinsic calibration information provided in the dataset, the transformation matrix $T_{lc}$ between the vehicle coordinate system and the virtual vehicle-mounted LiDAR coordinate system can be obtained. Based on the above transformation relationships, the rigid transformation model from the roadside world coordinate system to the virtual vehicle-mounted LiDAR coordinate system via the vehicle coordinate system can be established as
\begin{equation}
    T_{lw} = T_{lc} \cdot T_{cw}
\label{eq:T_lw}
\end{equation}

\begin{equation}
\begin{bmatrix}
x_l^i \\
y_l^i \\
z_l^i \\
1
\end{bmatrix}
=
\begin{bmatrix}
R_{lw} & t_{lw} \\
0 & 1
\end{bmatrix}
\begin{bmatrix}
x_w^i \\
y_w^i \\
z_w^i \\
1
\end{bmatrix}
\end{equation}
where $(x_w^i,y_w^i,z_w^i)^\top$ represents the 3D spatial position of point $p^i, i=1,\dots, N$ in the roadside world coordinate system, and $(x_l^i,y_l^i,z_l^i)^\top$ represents its 3D coordinate representation in the target vehicle-mounted LiDAR coordinate system.

After spatial alignment is completed, perception range constraints are applied to the point cloud data according to the effective detection distance of the virtual vehicle-mounted LiDAR. Let $r_i$ denote the radial distance from point $p_l^i$ to the LiDAR origin. The point is retained if $r_{min} \le r_i \le r_{max}$; otherwise, it is removed. After filtering, only the point cloud data within the effective perception range are retained, resulting in the point set $P_l=\{p_l^i\}_{i=1}^M$ in the LiDAR coordinate system, which is used for subsequent virtual scanning modeling and point cloud generation.

\subsection{Ground Scene Decomposition Module}

After completing cross-view spatial alignment and perception region filtering, the point cloud set in the vehicle-mounted LiDAR coordinate system is obtained as
$P_l=\{p_l^i\}_{i=1}^M$.
To support subsequent hierarchical modeling and virtual scan generation based on scene structure, it is first necessary to decompose the scene into ground and non-ground components. Specifically, the goal of ground segmentation is to divide the point set into a ground point set $P_{lg}$ and a non-ground point set $P_{ln}$, satisfying

\begin{equation}
P_l = P_{lg} \cup P_{ln}, \quad P_{lg} \cap P_{ln} = \emptyset.
\label{eq:p_lg_ln}
\end{equation}

To efficiently process large-scale point cloud data, this work adopts Patchwork++ \cite{lee2022patchwork++} as the ground segmentation preprocessing module. Patchwork++ is an efficient ground segmentation method designed for large-scale LiDAR point clouds. It achieves robust ground extraction in complex urban environments by combining polar coordinate space partitioning with local geometric consistency evaluation.

In the Patchwork++ framework, the point cloud is first mapped into polar coordinate space and divided into multiple local regions. For each local region, principal component analysis (PCA) is used to estimate the local plane normal vector and structural features, and ground classification is performed based on ground geometric consistency metrics.

In the proposed method, Patchwork++ is mainly used as a geometric pre-segmentation module to obtain the ground point set $P_{lg}$ and the non-ground point set $P_{ln}$. The decomposition results serve as inputs to the subsequent virtual scan point cloud generation module, where ground and non-ground point clouds are resampled using different geometric modeling and ray-casting strategies, respectively.

\subsection{Point Cloud Resampling Generation Module}

\begin{figure}[h]
    \centering
    \includegraphics[width=0.9\textwidth]{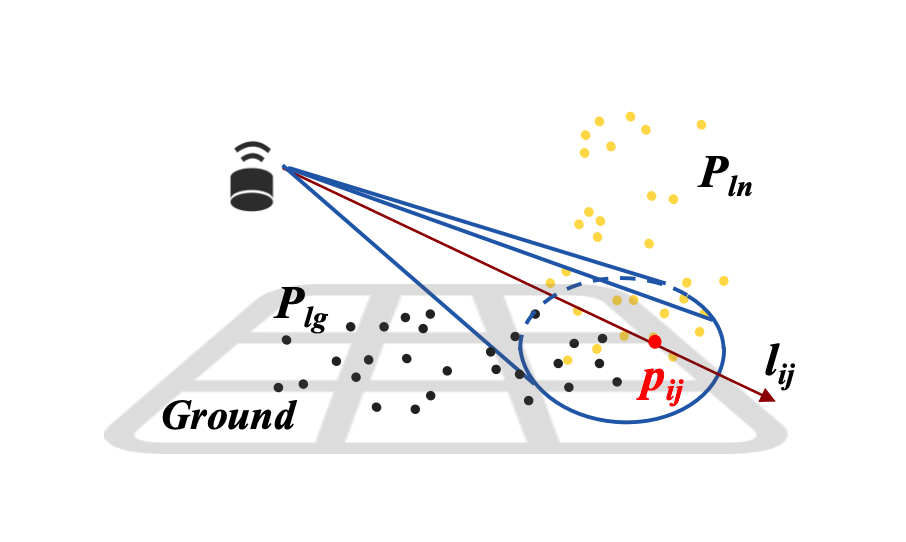}
    \caption{Schematic Representation of resampling point cloud for virtual LiDAR Ray $l_{ij}$. Resampling is first attempted from non-ground points $P_{ln}$. Only when the virtual ray receives no non-ground returns and is not occluded by non-ground objects, resampling is performed from ground points $P_{lg}$.}
    \label{fig4}
\end{figure}

After obtaining the non-ground point set $P_{ln}$ and ground point set $P_{lg}$, this module performs geometric modeling and ray-casting computation for the two types of point clouds based on the virtual LiDAR scanning model, in order to generate point cloud data under the target vehicle viewpoint, as illustrated in Fig. \ref{fig4}.
 
\textbf{Non-Ground Point Cloud Resampling:}  
For each point $p_n$ in the non-ground point set $P_{ln}$, it is first assigned to the corresponding virtual ray according to its polar angle $\theta_n$ and azimuth angle $\phi_n$. Points satisfying
$\theta_j \le \theta < \theta_{j+1} \quad \text{and} \quad \phi_i \le \phi < \phi_{i+1}$
are assigned to the virtual ray $l_{ij}$. After assignment, the non-ground point subset corresponding to virtual ray $l_{ij}$ is denoted as $P_{ij} \subseteq P_{ln}$.

For each non-empty point subset $P_{ij}$, the point closest to the LiDAR origin, denoted as $p_{min}$, is selected. Then, a subset of points $\{(x_k,y_k,z_k)\}_{k=1}^n$ whose distances to $p_{min}$ are smaller than a threshold $\sigma=1(\text{m})$ is selected. Based on these points, a local plane is fitted using the least squares method. The local plane is modeled in the following functional form:
\begin{equation}
z = ax + by + c
\end{equation}

The plane parameters are obtained by solving the following least squares problem:
\begin{equation}
\min_{a,b,c}
\sum_{k=1}^{n}
\left(z_k-(ax_k+by_k+c)\right)^2
\end{equation}

After obtaining the plane parameters, the ray parametric equation can be expressed as:
\begin{equation}
\mathbf{p}(t)=\mathbf{o}+t\mathbf{d}_{ij}
\end{equation}
where the unit direction vector of the virtual ray $\mathbf{d}_{ij}$ is defined in Eq.(\ref{eq:ray_direction}), and the LiDAR origin is $\mathbf{o}=(0,0,0)^\top$.

Using $\mathbf{n}=(a,b,-1)^\top$ to represent the normal vector of the fitted local plane, the intersection parameter between the ray and the plane can be expressed as:
\begin{equation}
t_0=
\frac{-c-\mathbf{n}^T\mathbf{o}}
{\mathbf{n}^T\mathbf{d}_{ij}}
\end{equation}

An intersection is considered valid only when $t_0 > 0$. The intersection point is then given by:
\begin{equation}
p_{ij}=\mathbf{o}+t_0\mathbf{d}_{ij}
\end{equation}

All valid intersection points are aggregated to form the non-ground point cloud set under the target vehicle viewpoint:
$V_n=\{p_{ij}\}$.

\textbf{Ground Point Cloud Resampling:}  
For the ground point cloud $P_{lg}$, this work assumes that the road surface in the intersection scene is approximately planar. Based on this assumption, a global least squares plane fitting is performed on the ground point cloud to obtain the ground plane model $S_{\text{ground}}$.

To simulate the occlusion effect of non-ground objects on the ground surface, virtual rays are grouped into spatial sectors. In the vertical direction, every two adjacent rays form one group, and in the horizontal direction, every 25 adjacent rays form one group, resulting in a sector. If non-ground returns exist within a sector, all rays in that sector are removed.

For the remaining rays, their intersections with the ground plane $S_{\text{ground}}$ are computed, and all valid intersection points are aggregated to form the ground point cloud set $V_g$.

For each resampled point, its reflection intensity is assigned using the intensity value of the closest original point to the LiDAR origin within the corresponding point subset.

Finally, the two types of point clouds are fused to obtain the complete virtual vehicle-mounted LiDAR point cloud:
\begin{equation}
V=V_n\cup V_g.
\end{equation}

\subsection{Cross-Domain Label Mapping Module}

After completing the target viewpoint point cloud resampling generation, it is necessary to simultaneously perform cross-domain mapping of annotation information from the roadside world coordinate system to the vehicle-mounted LiDAR coordinate system.

Based on the rigid transformation model in Eq.(\ref{eq:T_lw}), coordinate transformation of bounding box pose information can be achieved. During this process, object category information and geometric dimensions remain unchanged, while spatial position and orientation require coordinate transformation.

Let the object center coordinate in the roadside world coordinate system be $X_w$, and the object center coordinate in the vehicle-mounted LiDAR coordinate system be $X_l$. Then,
\begin{equation}
X_l = R_{lw} \cdot X_w + t_{lw}, \quad X_l \in \mathbb{R}^3
\label{eq:X_l_transform}
\end{equation}

Let the object rotation vector in the roadside world coordinate system be $\Theta_w$. Then, its pose in the vehicle-mounted LiDAR coordinate system can be expressed as:
\begin{equation}
\Theta_l = \mathrm{InvRodrigues}\left(R_{lw} \cdot \mathrm{Rodrigues}(\Theta_w)\right), 
\quad \Theta_l \in \mathbb{R}^3
\label{eq:theta_l_transform}
\end{equation}

Here, $X_l$ represents the 3D spatial position of the target object in the vehicle-mounted LiDAR coordinate system, and $\Theta_l$ represents the pose parameters of the target object in the vehicle-mounted LiDAR coordinate system, expressed in Euler angles $(\mathrm{roll}, \mathrm{pitch}, \mathrm{yaw})$.

Through the above transformation, roadside annotation information can be directly mapped to the target vehicle-mounted LiDAR viewpoint without re-annotation, thereby enabling cross-platform reuse of annotated data.

\section{Experimental Results}\label{sec4}

\subsection{Setup} 

This study utilizes our R2V-LiDAR dataset which comprises two subsets: a roadside dataset and a vehicle-mounted dataset. The roadside dataset was acquired using four 128-beam LiDARs deployed at an intersection, covering 89 time intervals including morning peaks, evening peaks, and off-peak hours. It contains over 10,000 frames in total, of which 1,293 frames when roadside and vehicle-side detection range are overlapped have been annotated. The annotation information includes object categories, 3D geometric dimensions, centroid coordinates, rotation vectors, and unique identifiers within continuous clips. The vehicle-mounted dataset was obtained by a data collection vehicle equipped with a Pandar64 LiDAR traversing the identical intersection, serving as the ground truth benchmark for comparison with the generated data.

Based on this dataset, and utilizing the collection vehicle as the target perspective, 1,122 frames of annotated target-view point cloud data were generated from the roadside dataset using the RS2AD-LiDAR system. Frames without sufficient overlap are excluded to ensure reliable geometric reconstruction. The effectiveness of the generated data is subsequently validated across two dimensions: point cloud similarity and object detection.

The key parameters of the virtual LiDAR employed in the experiments are configured in accordance with the actual specifications of the Pandar64 LiDAR, as detailed in Table \ref{tab:virtual_lidar_params}. The elevation angle distribution of its 64 scanning rays adheres to the sensor's factory calibration values, exhibiting a non-uniform distribution where channels in the middle and lower regions are denser to optimize the detection of near-ground targets.

\begin{table}[h]
\caption{Default parameters of the virtual LiDAR}
\label{tab:virtual_lidar_params}
\centering
\begin{tabular}{@{}llll@{}}
\toprule
Parameter & Symbol & Reference Value & Remarks \\
\midrule
Vertical FOV & $\alpha$   & $-25^{\circ} \sim 15^{\circ}$ & $0^{\circ}$ at horizon, positive upward \\
Horizontal FOV & $\beta$   & $360^{\circ}$ & -- \\
Number of beams & $k$   & 64 & -- \\
Horizontal angular resolution & $\Delta\beta$  & $0.2^{\circ}$  & -- \\
Maximum detection range & $r_{\max}$  & $200\,\mathrm{m}$  & -- \\
\bottomrule
\end{tabular}
\footnotetext{The parameters are configured according to the Pandar64 LiDAR used in the real vehicle-mounted dataset.}
\end{table}

\subsection{Semantic Similarity Evaluation of Generation Point Cloud}

To evaluate the semantic distribution consistency between the generated and real point clouds, we employ the PointTransformer model \cite{wu2024ptv3}, pretrained on the nuScenes dataset (without fine-tuning), to perform semantic segmentation on both datasets separately. The overall proportion of points in each semantic category is then compared. Given that the dataset used in this work lacks semantic annotation information, this experiment aims to directly verify the plausibility of the generated point clouds from the perspective of data distribution using the pretrained model.

\begin{table}[h]
\caption{Semantic distribution comparison between real and generated point clouds.}
\label{tab:semantic_comparison_table}
\centering
\begin{tabular}{@{}llll@{}}
\toprule
Label & Real Data (\%) & Generated Data (\%) & Diff (\%) \\
\midrule
barrier & 0.52 & 1.32 & +0.80 \\
bicycle & 0.49 & 0.45 & -0.04 \\
bus & 0.02 & 0.03 & +0.01 \\
car & 0.00 & 0.00 & 0.00 \\
construction\_vehicle & 0.01 & 0.01 & 0.00 \\
motorcycle & 0.19 & 0.16 & -0.03 \\
pedestrian & 0.08 & 0.00 & -0.08 \\
traffic\_cone & 7.42 & 8.39 & +0.97 \\
trailer & 0.28 & 0.32 & +0.04 \\
truck & 9.14 & 2.40 & -6.74 \\
driveable\_surface & 7.83 & 8.15 & +0.32 \\
other\_flat & 0.06 & 0.08 & +0.02 \\
sidewalk & 39.51 & 41.36 & +1.85 \\
terrain & 2.62 & 1.56 & -1.06 \\
manmade & 11.95 & 10.16 & -1.79 \\
vegetation & 28.88 & 25.61 & -3.27 \\
\midrule
\multicolumn{4}{l}{\textbf{Overall distribution similarity metrics:}} \\
\midrule
JS Distance $\downarrow$ & 0.1322 & Cosine Similarity $\uparrow$ & 0.9674 \\
\bottomrule
\end{tabular}
\end{table}

Table \ref{tab:semantic_comparison_table} presents the proportions of points and the corresponding differences across 16 semantic categories for both real and generated point clouds. It is observed that the distribution of the generated point clouds is highly consistent with that of the real data for the majority of categories. A notable discrepancy is found in the proportion of the \texttt{truck} category, which is primarily attributed to the insufficient object detail in the generated point clouds and the recognition limitations of the pretrained model across different datasets. Constrained by the sparsity of roadside point clouds, the generated point clouds exhibit less shape completeness regarding object details compared to real point clouds, thereby affecting the semantic segmentation results.

Furthermore, it is important to note that since this experiment directly utilizes a model pretrained on the nuScenes dataset without fine-tuning for our specific scene, the model's generalization ability for certain categories is limited. For instance, the recognition rates for vehicle categories such as \texttt{bus}, \texttt{car}, and \texttt{construction\_vehicle} are near zero. This anomaly is consistent across both real and generated point clouds, indicating that the discrepancy stems mainly from the generalization limitations of the model itself across datasets, rather than defects in the quality of the generated data. Nevertheless, from an overall perspective, the generated point clouds maintain distribution trends consistent with the real data for major scene elements such as \texttt{drivable\_surface}, \texttt{sidewalk}, and \texttt{vegetation}.

To further quantify the overall distribution similarity, two distribution similarity metrics, Jensen-Shannon distance and Cosine similarity, are calculated.

\begin{equation}
D_{\text{JS}}(P \parallel Q) = \sqrt{\frac{1}{2} D_{\text{KL}}(P \parallel M) + \frac{1}{2} D_{\text{KL}}(Q \parallel M)}
\end{equation}

\begin{equation}
S_{\text{cos}}(P, Q) = \frac{P \cdot Q}{|P| |Q|}
\end{equation}
\vspace{0.2cm}

Here, $P$ and $Q$ denote the semantic class probability distributions of the generated point cloud and the real point cloud, respectively, and $M = \frac{1}{2}(P+Q)$. $D_{\text{KL}}$ represents the Kullback-Leibler divergence. The Jensen–Shannon (JS) distance is defined as the square root of the JS divergence and is used to measure the similarity between two probability distributions. Its value lies within the interval $[0,1]$, where values closer to 0 indicate higher similarity between the two distributions. Cosine similarity measures the overall similarity between distributions by computing the cosine of the angle between two vectors in a high-dimensional space. Its value ranges from $[-1, 1]$, where values closer to 1 indicate that the two vectors have more consistent directions, i.e., the distributions are more similar.

The calculation results show that the cosine similarity reaches 0.967, which is close to 1, and the JS distance is only 0.132, indicating that the generated point cloud and the real point cloud exhibit high consistency in semantic class distribution.

\begin{figure}[htbp]
    \centering
    \includegraphics[width=0.9\textwidth]{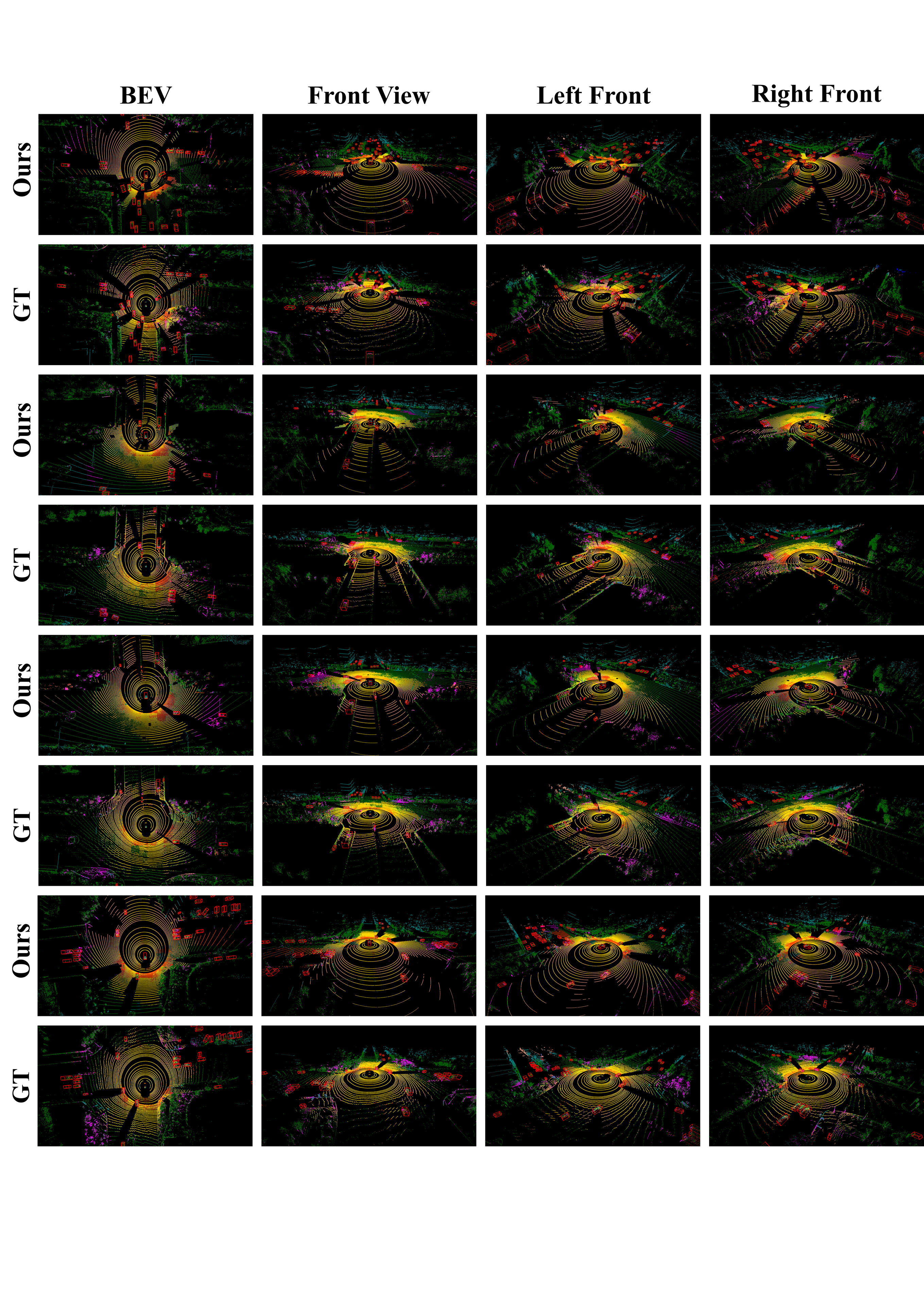}
    \caption{Qualitative comparison of semantic segmentation results between real and generated point clouds at nearly the same timestamp. The generated point cloud shows consistent macro-level semantic structures with the real data, particularly for classes such as drivable surface, sidewalk, and vegetation.}
    \label{fig6}
\end{figure}

Fig. \ref{fig6} illustrates a qualitative comparison of the semantic results for the real and generated point clouds at near timestamps (with a time difference of less than 0.1 s). It is evident that the two datasets generally align in macro semantic structures such as \texttt{drivable\_surface}, \texttt{sidewalk}, and \texttt{vegetation}, qualitatively demonstrating the rationality of the generated point clouds at the semantic level.

\subsection{Object Detection Experiments}

To validate the utility of the data generated by RS2AD-LiDAR for training in downstream autonomous driving tasks, a 3D object detection benchmark platform is constructed based on OpenPCDet \cite{openpcdet2020}. By incorporating the generated data as augmented samples into the training set, the impact on model performance is investigated.

The specific experimental setup is as follows: The control group utilizes only annotated real vehicle-mounted point cloud data to train a three-class object detection model; the experimental group incorporates point cloud data generated by RS2AD-LiDAR into the same real data for joint training. No other data augmentation strategies are applied during the training process. Both groups of models are evaluated on the same real data test set to ensure a fair comparison. The detection ranges are uniformly set to $x \in [-50, 50]$, $y \in [-40, 40]$, $z \in [-3, 1]$ (unit: meters). However, for the PointPillar \cite{lang2019pointpillars} model, due to the specific requirements of its network structure regarding the size ratio of the input voxel grid, the detection range is adjusted to $x \in [-51.2, 51.2]$, $y \in [-40.96, 40.96]$, $z \in [-3, 1]$ (unit: meters) to satisfy the stride design of its backbone network.

The evaluation metric adopts strict Average Precision (AP), specifically AP\_R40 \cite{simonelli2022disentangling}, which is computed at both the bird’s eye view (BEV) level and the 3D detection level.  
Detection performance is determined based on the Intersection over Union (IoU) between predicted bounding boxes and ground-truth bounding boxes, defined as:
\begin{equation}
\mathrm{IoU}=\frac{|B_{\text{pred}} \cap B_{\text{gt}}|}
{|B_{\text{pred}} \cup B_{\text{gt}}|}
\end{equation}
where $B_{\text{pred}}$ denotes the predicted bounding box and $B_{\text{gt}}$ denotes the ground-truth annotated bounding box. The IoU thresholds for each category follow the common practice in autonomous driving object detection, set to 0.7 for \texttt{Car} and 0.5 for \texttt{Pedestrian} and \texttt{Non\_motor\_rider}.

The dataset is split based on scene clips. The number ratio of clips in the training, validation, and test sets is $61:10:18$, and the corresponding frame ratio is approximately $7:1:2$. The validation set is used for selecting the optimal model weights, while the test set is used for final performance reporting.

The experiments cover three mainstream detection models: PointPillar \cite{lang2019pointpillars}, PV-RCNN \cite{shi2020pv}, and SECOND \cite{yan2018second}. The results are presented in Table \ref{tab:object_detection_results}.

\begin{table}
\centering
\caption{Evaluation Performance Under Different Training Data Configurations}
\label{tab:object_detection_results}
\begin{tabular}{lcccccccc}
\hline
\multirow{2}{*}{Model} & \multirow{2}{*}{Trainset} & \multicolumn{2}{c}{Car$_{\mathrm{IoU}=0.7}$} & \multicolumn{2}{c}{Pedestrian$_{\mathrm{IoU}=0.5}$} & \multicolumn{2}{c}{Non\_Motor\_Rider$_{\mathrm{IoU}=0.5}$} \\
\cmidrule(lr){3-8}
 & & BEV AP& 3D AP & BEV AP& 3D AP & BEV AP& 3D AP \\
\hline
\multirow{2}{*}{PointPillars} 
 & Real\footnotemark[1] & \textbf{88.13} & 69.02 & 24.68 & 3.44 & 82.29 & 75.02 \\
 & Mix\footnotemark[2]  & 88.11 & \textbf{69.85} & \textbf{25.32} & \textbf{5.01} & \textbf{82.40} & \textbf{75.48} \\
\hline
\multirow{2}{*}{SECOND} 
 & Real & \textbf{88.80} & 72.57 & 42.94 & 9.85 & \textbf{84.93} & 78.9901 \\
 & Mix  & 88.51 & \textbf{73.02} & \textbf{43.89} & \textbf{10.44} & 82.65 & \textbf{79.29} \\
\hline
\multirow{2}{*}{PV-RCNN} 
 & Real & 89.21 & 74.22 & 42.59 & 9.17 & \textbf{85.71} & \textbf{81.97} \\
 & Mix  & \textbf{89.29} & \textbf{77.06} & \textbf{46.16} & \textbf{11.92} & 85.30 & 81.48 \\
\hline
\end{tabular}
\footnotetext[1]{Real: Control group trained solely on real-world vehicle-mounted LiDAR point cloud data.}
\footnotetext[2]{Mix: Experimental group trained on the same real data with additional point clouds generated by RS2AD-LiDAR.}
\footnotetext{Bold values indicate better performance between the two training configurations for the same model.}
\end{table}

Experimental results indicate that incorporating generated data yields marginal but consistent improvements across most detection metrics. The data augmentation effect is particularly significant for the \texttt{Pedestrian} category, which is very scarce in the original dataset. All three models achieve performance gains in both BEV and 3D detection for this category. Specifically, the PV-RCNN model achieves an improvement of 3.57\% in BEV detection and 2.75\% in 3D detection for the \texttt{Pedestrian} category. For the other two categories, the detection results after adding generated data also generally exhibit an upward trend; for instance, the PV-RCNN model shows a 2.84\% improvement in 3D detection for the \texttt{Car} category, and the PointPillar model shows a 0.46\% improvement in 3D detection accuracy for the \texttt{Non\_motor\_rider} category. These changes in metrics demonstrate that when generated data is added to the training set as supplementary data, it effectively enhances the detection accuracy of the models in real-world test scenarios.

Admittedly, certain fluctuations are observed in the experimental results. In a few cases, the performance of the experimental group exhibits a slight decline. This may be attributed to subtle discrepancies in distribution between the generated and real data in areas deviating from the center of the intersection or in occluded scenarios. However, the magnitude of most fluctuations remains below 1\%, with no significant performance degradation, indicating that the generated data is stably compatible with real data.

A notable discrepancy is observed in the SECOND model for BEV detection of the \texttt{Non\_motor\_rider} category, which is 2.28\% lower than that of the control group. We hypothesize that this stems from the model's sensitivity to data distribution. The SECOND model employs a voxel-based sparse convolution backbone, which is relatively sensitive to the spatial distribution and density of the input point cloud. For targets with complex structures and sparse point clouds, such as \texttt{Non\_motor\_rider}, systematic deviations exist in the geometric details and spatial distribution of the generated data compared to real data (e.g., missing details like handlebars). Since BEV detection relies heavily on the precision of the object's bottom contour, this deviation is amplified in the BEV perspective.

Overall, the experimental results demonstrate that RS2AD-LiDAR generated data, acting as augmented samples, effectively enhances model performance, particularly for the long-tail \texttt{Pedestrian} category. Although the absolute improvement is limited, considering that this performance gain is achieved without any additional cost for real data collection and annotation, it fully substantiates the practical value of RS2AD-LiDAR generated data as an efficient, low-cost tool for data augmentation, holding significant value for the expansion of vehicle-mounted training data.

\begin{figure}[htbp]
    \centering
    \includegraphics[width=0.9\textwidth]{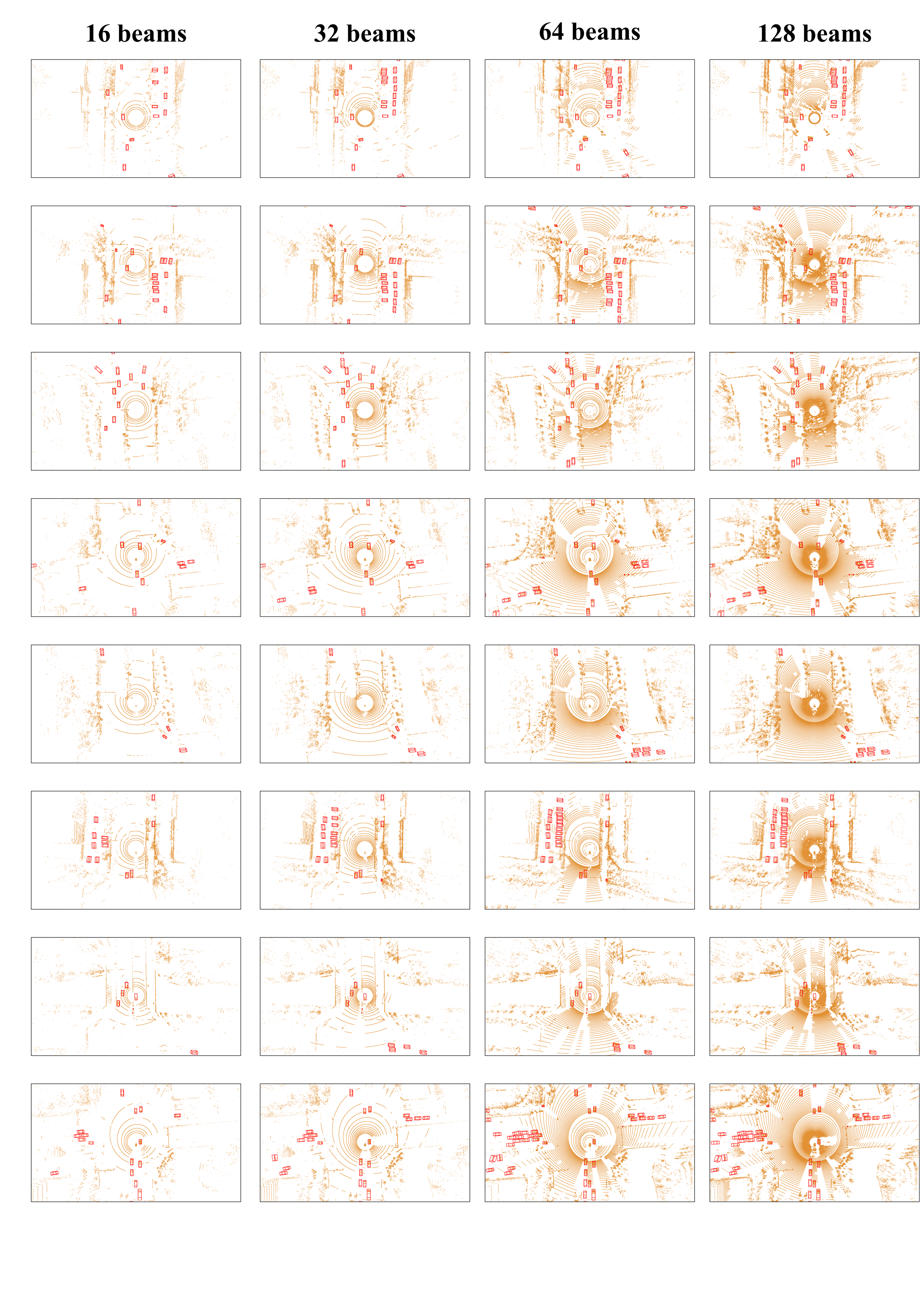}
    \caption{Qualitative comparison of generated vehicle-mounted point clouds under different virtual LiDAR configurations in the same roadside scene. The results include varying scan beam numbers: 16, 32, 64, and 128 beams. The comparison demonstrates the configurability of the proposed method.}
    \label{fig7}
\end{figure}

\subsection{Qualitative Demonstration of LiDAR Configurability}

In addition to utilizing the actual data collection vehicle as the target perspective, other annotated vehicles within the scene can also be selected as target perspectives. The resulting generated data is illustrated in Fig. \ref{fig2} upper part. Regarding the transformation matrix between the vehicle-mounted LiDAR and the vehicle body coordinate system: for the actual data collection vehicle, it is derived from the calibration relationship between the LiDAR and the vehicle 
bounding box center. For other vehicles in the scene, the virtual LiDAR is uniformly positioned 0.25 meters directly above the vehicle center in this demonstration; however, this placement can be flexibly adjusted according to specific requirements.

To validate the flexibility and configurability of the proposed virtual LiDAR model, we further generated point cloud results with multiple configurations by modifying key parameters under the same roadside scene. Fig. \ref{fig7} presents a comparison of the generated results under different numbers of scan beams (16 beams, 32 beams, 64 beams, and 128 beams). It can be observed that as the number of beams increases, the point cloud density becomes higher, and the representation of object surface details becomes richer.

It can be observed that a higher number of beams results in increased point cloud density and a richer characterization of object surface details. These results intuitively demonstrate that the proposed method is capable of flexibly adapting to different LiDAR model configurations by adjusting parameters, rather than being limited to the default 64-beam setting used in the experiments. This establishes a technical foundation for utilizing roadside data to generate multi-source and multi-configuration vehicle-mounted point clouds, thereby facilitating future research on sensor generalization and data diversity enhancement.

\begin{figure}[htbp]
    \centering
    \includegraphics[width=0.8\textwidth]{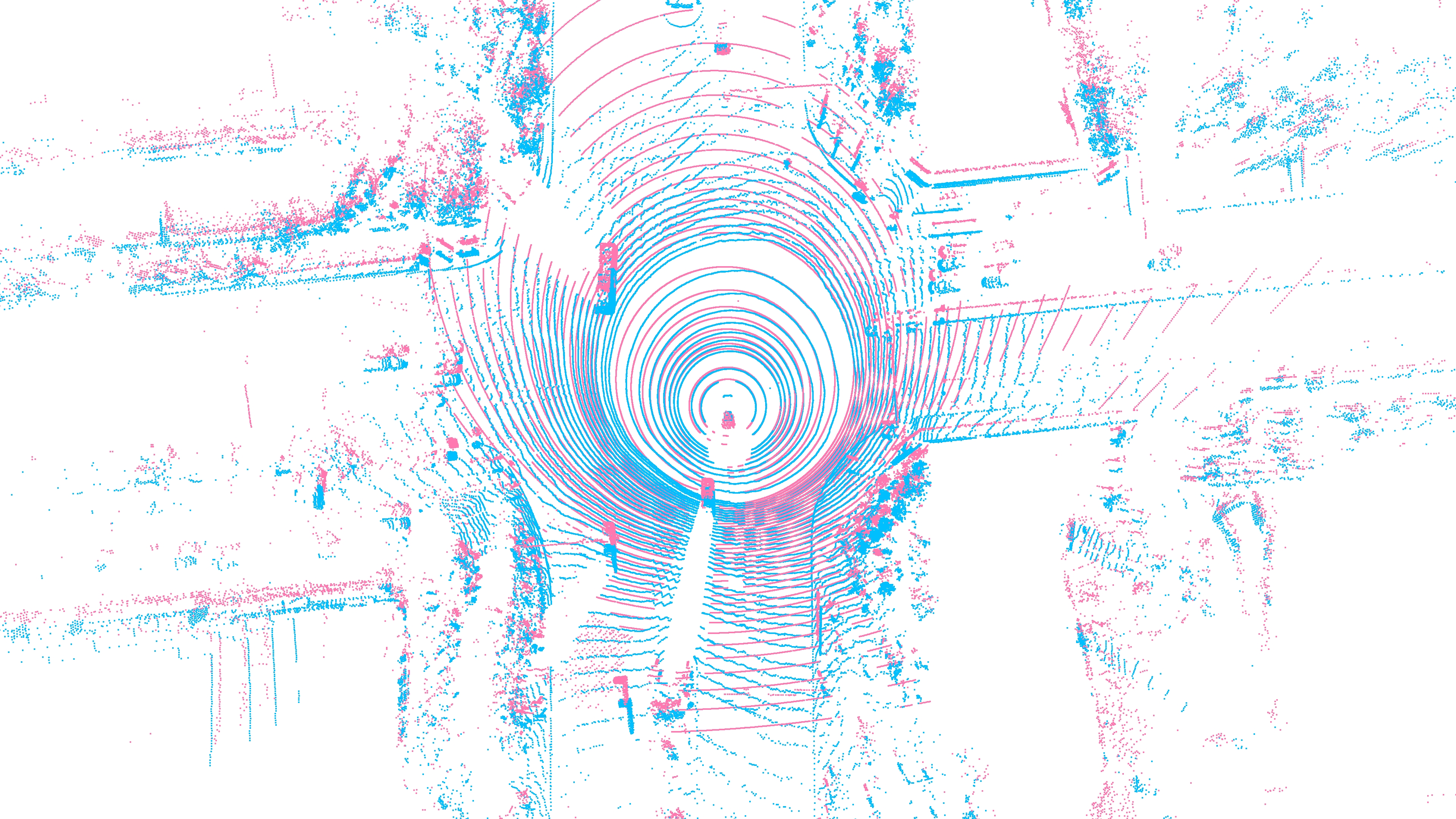}
    \caption{Spatial displacement caused by temporal asynchrony. The figure compares the generated vehicle-mounted point cloud(pink) with its temporally nearest real vehicle-mounted point cloud(blue), showing noticeable spatial misalignment due to the lack of precisely synchronized timestamps between roadside and vehicle-mounted sensors.}
    \label{fig5}
\end{figure}

Nevertheless, it is important to acknowledge a practical limitation inherent to the current experimental setting. Due to the temporal asynchrony between the roadside sensors and the vehicle-mounted LiDAR systems, there is no ground-truth vehicle-mounted point cloud that shares an exactly identical timestamp with the generated data. As illustrated in Fig.\ref{fig5}, even the temporally nearest real vehicle-mounted point cloud exhibits noticeable spatial displacement relative to the generated result.
Such misalignment primarily arises from unavoidable time offsets between heterogeneous sensing platforms, rather than deficiencies in the geometric modeling process itself. While this limitation does not affect the overall feasibility of the proposed framework, it inevitably introduces uncertainty in direct point-wise spatial comparison and quantitative alignment assessment, and therefore constitutes an important direction for future improvement.

\section{Conclusion}\label{sec5}

In this paper, we primarily focus on the key challenges faced in autonomous driving training data acquisition, including the high cost of data collection, the scarcity of high-value vehicle-side data, and the large discrepancies in cross-platform data representation. These challenges make it difficult for existing methods to efficiently scale training data while maintaining data realism. To address this issue, we propose RS2AD-LiDAR, a roadside-to-vehicle LiDAR point cloud generation framework based on cross-view geometric modeling. By leveraging a virtual vehicle-mounted LiDAR scanning mechanism and introducing cross-coordinate spatial alignment and scene structure decomposition strategies, we effectively achieve geometric mapping and data reconstruction between the roadside observation space and the vehicle observation space, providing an efficient and low-cost solution for autonomous driving data construction.

Extensive experiments demonstrate the advantages of the proposed framework in terms of data realism and task usability. The experimental results show that the point clouds generated by RS2AD-LiDAR maintain high consistency with real vehicle-side point clouds in semantic distribution. In downstream task evaluation, using the generated data as training set augmentation samples effectively improves the detection performance of several mainstream object detection models on real test scenarios. In particular, more significant performance improvements are observed in rare categories (e.g., Pedestrian), demonstrating that the generated data possesses strong realism and training utility.

Although RS2AD-LiDAR performs well on the current self-collected intersection dataset, it is still limited by the time synchronization issues between roadside sensors and vehicle-mounted sensors, as well as the insufficient diversity of training data scenarios. Future work will be carried out in two directions. On the one hand, we will explore the introduction of time-aware generative models to mitigate the impact of temporal asynchrony among multi-source sensors, thereby further improving the spatiotemporal consistency and practical applicability of the generated data. On the other hand, we will expand data collection to more diverse traffic scenarios to verify the generalization capability of the proposed method and promote its application in more complex real-world environments.


\bibliography{sn-bibliography}

\end{document}